%% file: bare_jrnl.tex
\documentclass[journal]{IEEEtran}
\ifCLASSINFOpdf
\else
\fi
\usepackage{soul,xcolor,color}
\definecolor{orange}{RGB}{253,174,97}
\definecolor{red}{RGB}{215,25,28}
\definecolor{green}{RGB}{171,221,164}
\definecolor{blue}{RGB}{43,131,186}
\definecolor{yellow}{RGB}{255,255,191}

\usepackage{amsmath,amsfonts,amssymb}
\usepackage{amsmath}
\usepackage{subcaption}
\usepackage{graphicx}
\usepackage{paralist}
\usepackage{changepage}

\newcommand{\etal}[1]{#1~\textit{et al.}}
\newcommand{\U}{U-Net}
\newcommand{\model}{iW-Net}
\newcommand{\M}{$\mathcal{M}$}

\begin{document}
\bstctlcite{IEEEexample:BSTcontrol}
%\title{Bare Demo of IEEEtran.cls\\ for IEEE Journals}
\title{\model{}: an automatic and minimalistic interactive lung nodule segmentation deep network}

\author{Guilherme~Aresta$^{1,2,*}$,
        Colin~Jacobs$^{3}$,
        Teresa~Ara\'{u}jo$^{1,2}$,
        Ant\'{o}nio~Cunha$^{1,4}$,
        Isabel~Ramos$^{5}$,
        Bram~van~Ginneken$^{3}$ and
        Aur\'{e}lio~Campilho$^{1,2}$% <-this % stops a space
\thanks{$^{*}$ \small{\texttt{guilherme.m.aresta@inesctec.pt}}}% <-this % stops a space
\thanks{$^{1}$ INESC TEC -  Institute for Systems and Computer Engineering, Technology and Science, Rua Doutor Roberto Frias, 4200-465 Porto, Portugal}%
\thanks{$^{2}$Faculty of Engineering of University of Porto, Rua Doutor Roberto Frias, 4200-465 Porto, Portugal}%
\thanks{Radboud University Medical Center,6525 Nijmegen, The Netherlands}%
\thanks{$^{4}$University of Tr\'{a}s-os-Montes e Alto Douro, Quinta de Prados, 5001-801 Vila Real, Portugal}%
\thanks{$^{5}$Faculty of Medicine of University of Porto, Alameda Prof. Hern\^{a}ni Monteiro, 4200-319 Porto, Portugal}%
\thanks{Guilherme~Aresta is funded by the FCT grant SFRH/BD/120435/ 2016.
Teresa~Ara\'{u}jo is funded by the FCT grant SFRH/BD/122365/ 2016. 
This study is associated with project NLST-375 and LNDetector, which is financed by the ERDF - European Regional Development Fund through the Operational Programme for Competitiveness - COMPETE 2020 Programme and by the National Fundus through the Portuguese funding agency, FCT - Funda\c{c}\~{a}o para a Ci\^{e}ncia e Tecnologia within project POCI-01-0145-FEDER-016673.}%
}

\maketitle

\begin{abstract}

We propose \model{}, a deep learning model that allows for both automatic and interactive segmentation of lung nodules in computed tomography images.
\model{} is composed of two blocks: the first one provides an automatic segmentation and the second one allows to correct it by analyzing 2 points introduced by the user in the nodule's boundary. For this purpose, a physics inspired weight map that takes the user input into account is proposed, which is used both as a feature map and in the system's loss function.
Our approach is extensively evaluated on the public LIDC-IDRI dataset, where we achieve a state-of-the-art performance of 0.55 intersection over union \textit{vs} the 0.59 inter-observer agreement. Also, we show that \model{} allows to correct the segmentation of small nodules, essential for proper patient referral decision, as well as improve the segmentation of the challenging non-solid nodules and thus may be an important tool for increasing the early diagnosis of lung cancer.

\end{abstract}


% \fi
\IEEEpeerreviewmaketitle

\input{introduction}

\input{methods}

\input{results}

\input{conclusions}

% or


%\appendices

%\section{}

%\section*{Acknowledgments}

\ifCLASSOPTIONcaptionsoff
  \newpage
\fi



\bibliographystyle{IEEEtran}
\bibliography{Mendeley_LNDetector.NoduleDetection.bib}
\end{document}

%% file: introduction.tex
\section{Introduction}

Lung cancer is the most fatal cancer type in both men and women~\cite{Siegel2018}.
Thankfully, early diagnosis of this pathology and proper medical follow-up allow to increase the patients' survival rate. Namely, annual screening of risk groups with low-dose chest computed tomography (LDCT) allows to reduce lung cancer mortality by 20\%~\cite{doi:10.1056/NEJMoa1102873}.
During screening, radiologists search for lung nodules by visually inspecting the LDCT volumes. Potential findings are then characterized in terms of dimension (axes length and volume), texture (solid, sub-solid and non-solid), spiculation, calcification and location. Patient follow-up is then decided according to a specific lung cancer screening guideline.
Particularly, the initial nodule dimensions and growth-rate are two pivotal characteristics in major screening guidelines~\cite{ACR2014, CallisterMEJBaldwinDRAkramAR2015, MacMahon2017} and thus accurate 3D lung nodule segmentation is an important task during screening. However, performing accurate manual segmentation is a highly time consuming task, thus motivating the need for automatic lung nodule segmentation solutions.  Furthermore, it is known that nodule segmentation is a subjective task and specialists often disagree on their annotations~\cite{Armato2011}. Consequently, interactive segmentation tools are of high interest on this clinical setting. 
\par
Over the past years, several automatic lung nodule segmentation methods have been proposed with the goal of automating lung cancer screening. Despite achieving acceptable performances, lung nodule segmentation methods are still limited because either do not allow for user interaction, are slow or require extensive user interaction (\textit{e.g.} adjustment of several parameters) to achieve a satisfying result.

\begin{figure}
\centering
\includegraphics[width=0.86\columnwidth]{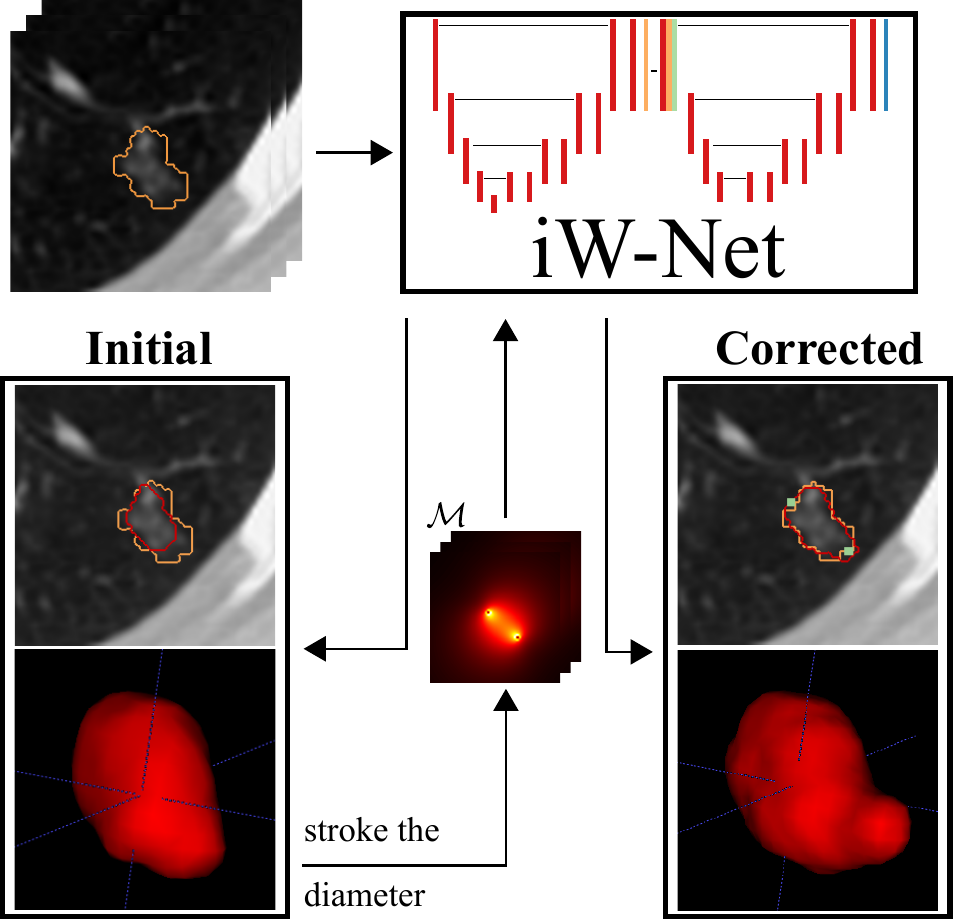}
\caption{Automatic and interactive lung nodule segmentations using \model{}.{\color{orange}$\blacksquare$}~ground-truth; {\color{red}$\blacksquare$}~prediction; {\color{green}$\blacksquare$}~end-points.  \label{fig:pipeline}}
\end{figure}

\subsection{Contributions}

We propose an end-to-end deep learning scheme, \model{} (interactive W-Net), that allows for both automatic and optional interactive 3D lung nodule segmentation, as suggested in Fig.~\ref{fig:pipeline}. The network receives as input a cube of fixed dimensions which centroid is indicated by the user, or by an automatic nodule detection framework, and proposes a corresponding segmentation. If the user is not satisfied, the segmentation can be corrected by using the end-points of a manually inserted stroke of the nodule's diameter. For this purpose, we use a second segmentation network that integrates the 3D image of the nodule, the initial segmentation and the coordinates of the end-points. Namely, this paper shows that the end-points can be represented by a physics-inspired weight map that, when used as a feature map and as loss function term, allows to cap the inter-observer agreement in the LIDC-IDRI public dataset.
%Specifically, the proposed method uses a physics-inspired weight map that affects the segmentation process by integrating two user-inserted points. 
Our approach allows a simple and fast segmentation correction when that information is available without introducing a significant over-head in comparison to the non-guided version of the model.

\input{state_art.tex}

%Brief state of the art review

%Motivation and contributions

%Interestingly, the smoothness term allows the model to improve the initial segmentations even when no scribbles are provided.
%However, the performance of the model is dependent on a set of extra hyperparameters, including the number of fine-tuning iterations
%Despite its high performance, 

\par~\par
Having in mind the limitations of the existing approaches for lung nodule segmentation, we propose \model{}, a simplistic deep learning approach that allows to alter segmentations while requiring minimal user interaction. The model's design and the respective experimental setup are described in Section~\ref{sec:methods}. Then, in Section~\ref{sec:results} we show that our approach allows to achieve state-of-the-art performance. Finally, Section~\ref{sec:conclusions} draws the main conclusions from this study.

%% file: state_art.tex
\subsection{Related works}

Lung nodule segmentation has been a focused research topic over the last decade.
Segmentation methods usually take advantage of the natural characteristics of solid nodules, which commonly have high contrast with the lung parenchyma and spherical shapes. A common approach is to do voxel-wise segmentation by extracting intensity~\cite{6825802,Tan2013b} and shape-related features, namely from Hessian matrices~\cite{Goncalves20161}, and training classifiers such as Support Vector Machines or Neural Networks~\cite{Messay201548} to obtain the final result. However, the extension of feature-design approaches for non-solid and sub-solid nodules is a hard and tedious process~\cite{Lassen2015} due to the cloudy texture, irregular shape and reduced contrast with the parenchyma of non-solid, and the diffused boundaries of sub-solid nodules.

Because of this, Convolutional neural networks (CNNs) have become the standard approach for medical image segmentation since they allow to significantly reduce the required field-knowledge to work with these images and thus the need for manual feature design. 
For instance, \etal{Wang}~\cite{Wang2017} proposed a multi-scale CNN that performs voxel-wise predictions, inside a cube containing a lung nodule, of the abnormal tissue. Each predicted voxel corresponds to the center of a fixed dimension patch to be processed by the network and thus predicting an entire segmentation requires the evaluation of a high number of patches. Furthermore, this model has an inherent lack of global context, since the network only evaluates patches, and thus the 3D reconstruction of the nodule may be affected.
A common solution is to adapt 3D \U{}~\cite{10.1007/978-3-319-46723-8_49} architectures, since they allow to consider both local and global context. With this in mind, \etal{Wu}~\cite{Wu2018} proposed a multi-task scheme for pulmonary nodule segmentation together with the prediction of the nodules' expected malignancy, achieving state-of-the-art performance in both tasks. This malignancy prediction is performed by concatenating and processing via a set of fully-connected layers the features of the segmentation network's bottle neck with a convolved version of the produced segmentation prediction.

Despite the high performance of deep learning methods, their application in the medical field is being criticized due to 
\begin{inparaenum}[1)]
\item the inherent lack of explanations behind the decision and,
\item the production of deterministic outputs, ignoring the existing inter-observer variability of the annotations and inhibiting the medical specialist to interact and change the decisions of the system.
\end{inparaenum}
With this in mind, \etal{Kohl}~\cite{Kohl2018} proposed to model the inter-observer variability by combining a conditional variational auto encoder (cVAE) with an U-Net. The cVAE is used for drawing a set of feature maps sampled from the trained latent space representation. These features are then concatenated with the last feature maps of the U-Net, which are then convolved to produce the segmentation output. By varying the sampled set of features from the cVAE, this model is capable of producing different, yet plausible, nodule segmentations. However, the method of \etal{Kohl} does not allow the clinician to alter the segmentation, instead forcing the specialist to opt for the result closer to his/her expectations.

Recently, \etal{Wang}~\cite{Wang2018} proposed a scribble-based approach to refine 2D and 3D segmentations resulting from a fully-convolutional neural network. First, the user selects a bounding box containing the anatomical structure to segment. 
For each unseen image, the top of a pre-trained segmentation model is trained to accommodate the foreground and background scribbles by minimizing, via an expectation-maximization (EM) approach, a loss function composed of two terms:
\begin{inparaenum}[1)]
\item a pixel-wise weighted categorical cross-entropy term that prioritizes the inclusion of foreground and the removal of background scribbles, and
\item a pair-wise smoothness term that encourages the aggregation of neighbor pixels of similar intensity~\cite{Rother2004}.
\end{inparaenum}
Even though this scheme achieves state-of-the-art results on organ segmentation in MRI images, its application for lung nodule segmentation is limited due to the nature of the abnormalities. For instance, nodules are often attached to structures of similar intensity, such as the pleural wall and blood vessels, and thus the EM scheme may lead to the inclusion of these structures in the segmentation and thus extra manual correction efforts. Also, sub-solid and non-solid nodules do not have a clear boundary, which can further hinder the minimization of the smoothness term.% of the loss function.

%% file: methods.tex
\section{Methods}
\label{sec:methods}

\begin{figure*}
\centering
\includegraphics[width=0.75\textwidth]{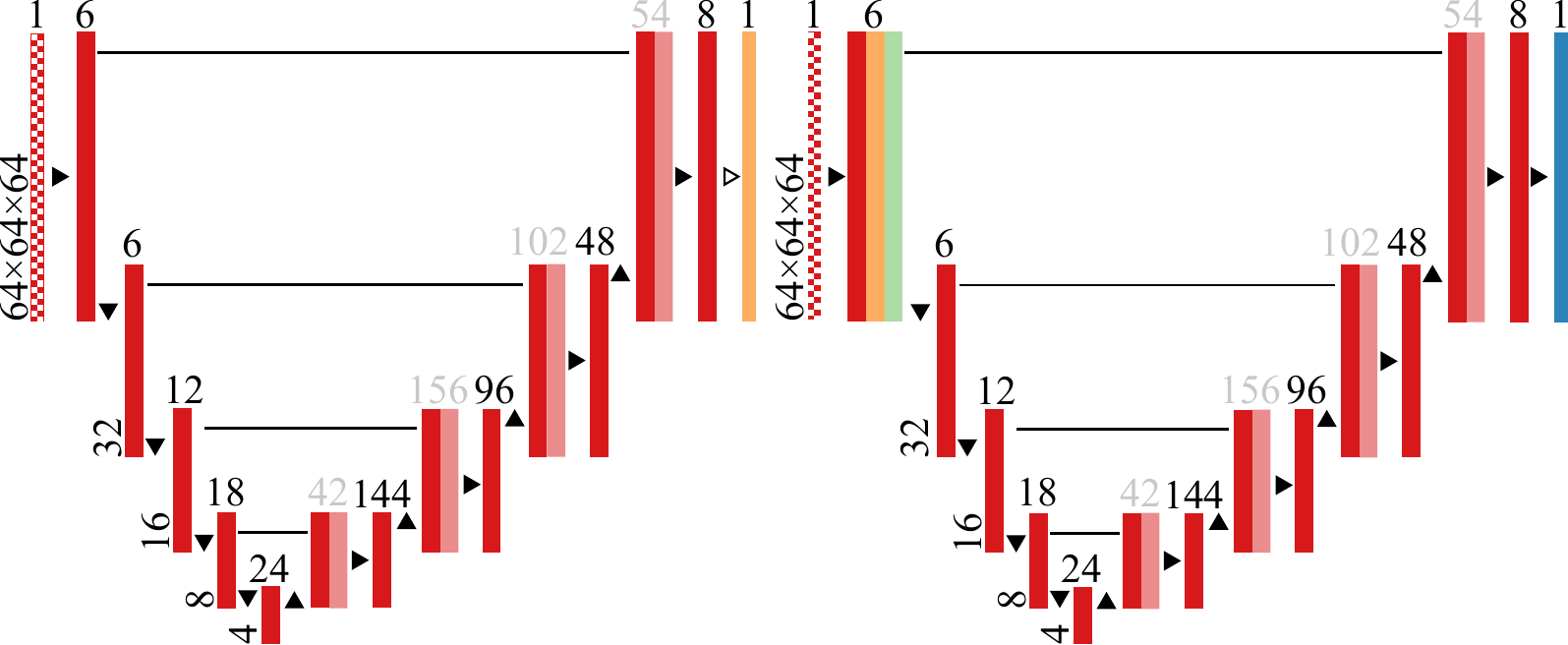}
\caption{\model: a network for guided segmentation of lung nodules, composed by a block responsible for predicting the initial segmentation and a second block for its correction. \protect\rotatebox{90}{$S$} is the side of the feature map. \protect\includegraphics[height=1.5ex]{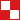}~input image {\color{red}$\blacksquare$} intermediary feature maps; {\color{orange}$\blacksquare$}~initial segmentation prediction; {\color{green}$\blacksquare$}~weight map \M{} computed from the user's input; {\color{blue}$\blacksquare$}~corrected segmentation. $\blacktriangleright$~$3\times 3 \times 3 \times N$ convolution, followed by batch normalization and rectified linear unit activation ($N$ is the number of feature maps, indicated on the top of each layer); $\blacktriangledown$~$3\times 3 \times 3 \times N$ convolution with stride $2\times 2 \times 2$, followed by batch normalization and rectified linear unit activation; $\blacktriangle$ $2 \times 2 \times 2$ nearest neighbor u-psample;$\triangleright$~$3\times 3 \times 3 \times N$ convolution with sigmoid activation. \label{fig:iwnet}}
\end{figure*}

\model{} allows to easily correct lung nodule segmentations according to the specialists' perception. As depicted in Fig.~\ref{fig:iwnet}, \model{} first performs an
\begin{inparaenum}
\item automatic 3D segmentation of lung nodules, predicted by the first block (\textit{i.e.} U) of the network, and after an
\item optional segmentation correction, performed by the second block via the analysis of the end-points of the user-introduced stroke of the diameter of the nodule.
\end{inparaenum}
For this, we propose a pixel-wise weight map \M{} to guide the segmentation, as detailed in Section~\ref{sec:me_map}. \M{} is then used as a feature map of \model{} and in a loss function term to train an auto-encoder segmentation network, as described in Sections~\ref{sec:me_unet}~and~~\ref{sec:me_loss}. 

\subsection{Weight map for segmentation control}
\label{sec:me_map}

Our weight map \M{} is inspired on the attraction field generated by punctual electric charges of opposite value.
Let $S$ define a sphere of undetermined radius:

\begin{equation}
S(x,y,z) = (x-x_0)^2+(y-y_0)^2+(z-z_0)^2
\end{equation}

\noindent where ($x_0$, $y_0$, $z_0$) is the center of the sphere and ($x_i$,$y_i$,$z_i$) are Cartesian coordinates. The unitary normalized gradient field is:% $\nabla S$, is defined as

\begin{equation}
\nabla S = \frac{2(x-x_0)+2(y-y_0)+(z-z_0)}{\sqrt{((2(x-x_0))^2+((2(y-y_0))^2+((2(z-z_0))^2}}
\end{equation}

\begin{figure}
\centering
\begin{subfigure}{0.35\columnwidth}
\includegraphics[width=\textwidth]{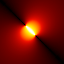}
\caption{$p=0$\label{fig:w0}}
\end{subfigure}
\quad
\begin{subfigure}{0.35\columnwidth}
\includegraphics[width=\textwidth]{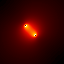}
\caption{$p=0.5$\label{fig:w0.5}}
\end{subfigure}

\begin{subfigure}{0.35\columnwidth}
\includegraphics[width=\textwidth]{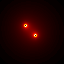}
\caption{$p=1$\label{fig:w1}}
\end{subfigure}
\quad
\begin{subfigure}{0.35\columnwidth}
\includegraphics[width=\textwidth]{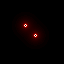}
\caption{$p=2$\label{fig:w2}}
\end{subfigure}

\caption{Examples of weight maps (middle slice is shown) with different decay values $p$. Colorbar: 0~\protect\includegraphics[width=4em,height=0.5em]{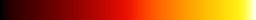}~1 \label{fig:weight_maps}}

\end{figure}

%The norm of the vectors of $\nabla S$ can be weighted as function of the distance to the center of the sphere resulting on a gradient mapping $Q$:
The norm of the vectors of $\nabla S$ can be weighted as function of the distance to the center of the sphere: %resulting on a gradient mapping $Q$:

\begin{equation}
Q_a = (-1)^a\frac{\nabla S}{|\nabla S|^p}
\end{equation}

\noindent where $p\in {\rm I\!R}$ controls the decay of the vectors' magnitude and $a\in\{0,\,1\}$ makes the field centripetal or centrifuge, respectively.
%For instance, for $p=2$, $Q$ corresponds to the electric field of a positive punctual charge and consequently 
Then, $W = Q_0+  Q_1$ is a vector field that moves from $Q_0$ to $Q_1$. In our approach, $Q_0$ and $Q_1$ correspond to the user introduced points and \M{}~$=|W|$ is a 3D feature map indicating how valuable each voxel is for the segmentation. In terms of magnitude, \M{} has high intensity in the region between the centers of $Q_0$ and $Q_1$ and low vector magnitude elsewhere, indicating to the network that the region between the two points has high interest for the segmentation. 
Changing $p$ affects the strength of the interaction between the two points, as shown in Fig.~\ref{fig:weight_maps}. Namely, a lower $p$ increases the the focus on the central region but also increases its overall volume, whereas a high $p$ leads to more spherical regions of interest surrounding the points. Note that if no points exist, then \M{} is a zero-value tensor with the same size of the input volume.

\subsection{\model{} for nodule segmentation}
\label{sec:me_unet}

The proposed nodule segmentation scheme is adaptation of the 3D U-Net~\cite{10.1007/978-3-319-46723-8_49}. As shown in Fig.~\ref{fig:iwnet}, \model{} is composed of two auto-encoders: the first outputs an initial segmentation, which is then used as an input for the second block to produce the corrected segmentation. Each of the auto-encoders has a reduced a number of filters in the encoding and decoding parts in comparison to the 3D U-Net, resulting in less parameters to tune and thus easing the back-propagation process.

\par
We include the proposed segmentation weight map \M{} by concatenating it to the initial feature maps of the encoding part of the second block of the model since preliminary experiments showed a significant performance drop if \M{} was included on the upsampling part only. In fact, adding \M{} on the initial part of segmentation correction block ensures that all weights of the model are affected by these external features. Due to the skip connections, \M{} is also included on the final segmentation layer, thus directly affecting the model's output. \par

\subsection{Loss function}
\label{sec:me_loss}

\model{} predicts a 3D map of the probability of each voxel belonging to the nodule. 
We use a two-term loss function, where the first is based on the intersection over union (IoU):

\begin{equation}
\mathcal{L}_{\text{IoU}} = 1 - \text{IoU} = 1 - \frac{\sum{I_t\circ I_p}}{\sum{(I_t+I_p)-\sum{I_t\circ I_p}}},
\label{eq:loss_iou}
\end{equation}
where $I_t$ and $I_p$ are the ground truth mask and the soft prediction mask, respectively, and ${\circ}$ is the Hadamard product. %In this work, $I_t$ is binary and $I_p$ is continuous in
The second term aims at forcing the network to have in account the manually introduced points by evaluating if there are segmentation points in the defined region of interest:

\begin{equation}
\mathcal{L}_{\text{attraction}} = 1 - \frac{\sum{((\mathcal{M}>\gamma)\circ I_p)}}{\sum{(\mathcal{M}>\gamma)}},
\label{eq:loss_attract}
\end{equation}
\noindent where $\gamma \in [0 \,1]$ controls the extent of the region of interest.  

The global loss $\mathcal{L}$ is the linear combination of Eqs.~\ref{eq:loss_iou}~and~\ref{eq:loss_attract}:
%Having in account that both $\mathcal{L}_{\text{IoU}}$ and $\mathcal{L}_{\text{attraction}}$ are in the range $[0 \,1]$, the global loss $\mathcal{L}$ is thus

\begin{equation}
\mathcal{L} = \lambda_1\mathcal{L}_{\text{IoU}} + (1-\lambda_1)\mathcal{L}_{\text{attraction}}
\label{eq:loss}
\end{equation}

\noindent where $\lambda_1$ controls the relative importance of the terms.

\subsection{Dataset and training details}

\model{} was developed using the LIDC-IDRI~\cite{Armato2011} dataset, which contains 1012 LDCT scans with variable slice thickness. In this dataset, nodules with diameter $\geq 3~mm$ have voxel-wise annotations from up to 4 different expert radiologists and the corresponding inter-observer agreement level is indicative of how likely an abnormality is in fact a nodule.
The dataset also contains a numeric description $\in \mathbb{N}$ of several nodule characteristics. Namely, nodule texture $\in [1,5]$ indicates the opacity of the nodule, with 1 being a pure non-solid nodule and 5 a pure solid nodule.
We considered the 888 scans used for the LUNA16 challenge~\cite{Setio2016} and studied 2284 nodules (some samples were discarded due to annotation inconsistencies, poor scan reconstruction or excessive slice thickness). From those, 1593, 1190 and 790 have agreement level $\geq 2$, $\geq 3$ and $\geq 4$, respectively.  In our experiments, a nodule is considered non-solid if it has an average texture $\leq 2$, solid if $=5$ and sub-solid otherwise. For an agreement level $\geq 2$, the dataset has 135 non-solid, 300 sub-solid and 1695 solid nodules. %385 nodules were discarded in relation to the initial LIDC-IDRI

All nodules were collected by patching a $51\times 51 \times 51~mm$ cube centered at the average center of mass of the specialists annotations and were then isotropically resized to $64\times 64 \times 64$~voxels. The intensity of the volume image was linearly mapped from $[-1000~400]$ Hounsfield Units to $[0~1]$. Adam~\cite{DBLP:journals/corr/KingmaB14} was used as optimizer (learning rate 0.001) and the network was trained using a batch size of 8 samples.

The dataset was artificially augmented by performing random rotations, translations, flips and zooms. For each epoch, user input was simulated by selecting the two most distant points on the middle axial slice of the segmentation. All agreement levels were considered to account for the inter-observer variability and thus no segmentation combination was performed, \textit{i.e.} the same nodule was paired with different viable ground-truths to train the model. Furthermore, \model{} was evaluated via stratified 5-fold cross-validation with partition at scan level and we used 20\% of the training for validation.
All hyper-parameters were found via random search~\cite{Bergstra2011} with 100 search steps. At each step, $\{\lambda_1,\gamma,p\} \sim U([0,1])$, where $U$ is an uniform distribution.  
Optimization was performed on the validation set of the first train-test split.
%The parameters that  maximized the IoU on the validation set of the first train-test split were kept.  
%1540+3415+864 = 5819

\model{} was trained in two steps. The first block was initially trained separately using $\mathcal{L}_{\text{IoU}}$ until the validation loss stopped improving for 3 epochs. The weights were then frozen and the entire \model{} was trained using $\mathcal{L}$, the output of the first segmentation block and the artificially generated user interaction until the loss stopped improving for 5 epochs. Since each nodule can have multiple segmentations (one per expert), \model{} had to perform different corrections according to the expert's annotation and the respective simulated user input. 
Experiments were performed on an Intel Core i7-5960X, 32Gb RAM, $2\times$GTX1080 desktop with Python 3.5 and Keras 2.2 \footnote{https://github.com/gmaresta/iW-Net}.

\subsection{Experiments and evaluation}

\model{} produces pixel-wise predictions $\in [0~1]$, which are thresholded at 0.5 for the model's evaluation.
The predictions are evaluated in terms of 3D intersection over union (IoU) and average surface distance (ASD), as follows:
%\footnote{computed using https://github.com/deepmind/surface-distance})
\begin{equation}
\text{IoU}(S,\hat{S}) = \frac{(S \cap \hat{S})}{(S \cup \hat{S})}
\end{equation}

\begin{align}
\text{ASD}(S,\hat{S}) &= \frac{1}{2}\Bigg( \frac{1}{N_S}\sum_i^{N_{S}}  \text{min}\left( d\left(S_i,\hat{S} \right) \right)  \\ 
& + \frac{1}{N_{\hat{S}}}\sum_i^{N_{\hat{S}}}  \text{min}\left( d\left(\hat{S}_i, S \right) \right) \Bigg) \nonumber
\end{align}

\noindent where $S$ is the expert's annotation, $\hat{S}$ is the model's prediction, $N_S$ and $N_{\hat{S}}$ are the number of surface elements, $d$ is the Euclidean distance (mm) and $\text{min}$ is the minimum operation. 

For each nodule, the average inter-observer IoU performance is computed by iteratively considering one expert's annotation as the ground-truth and the remaining as predictions and then averaging the results. For instance, the inter-observer IoU performance in an agreement level 4 nodule is the average of $12=4\, \text{annotators}\times 3\,\text{predictions}$ IoU results. For better comparison with the observers,
\model{} is only evaluated in nodules with agreement level $\geq 2$. 
The segmentation performance is also analyzed in terms of nodule radius and texture. 
We consider the radius of each nodule as the average of the equivalent spherical radius of all the annotators.

\paragraph{Experiment 1}
We study the performance of the non-guided segmentation unit (the first block of \model) using as comparison terms the average inter-observer agreement and the segmentation produced using the 3D~\U{}~\cite{10.1007/978-3-319-46723-8_49}. This \U{} is trained and tested on the aforementioned dataset. Due to computational constraints, the batch size is reduced to 2.
Evaluation is performed according to Eq.~\ref{eq:eval_1}:

\begin{equation}
\overline{\text{IoU}}\textbf{} = \frac{1}{N} \sum_{n=1}^N \text{IoU}(S_{n,j},\hat{S}_j)
\label{eq:eval_1}
\end{equation}

\noindent where $N$ is the expert's agreement level for nodule $j$. Since a nodule can have multiple segmentations, it is not expected that the model outperforms the inter-observer agreement.

\paragraph{Experiment 2}

The goal of this experiment is to evaluate the impact of the user's input on the segmentation of \model. For that, we artificially generate user inputs on the axial plane of the slice that contains the nodule's centroid. Similarly to the training procedure, the two most points distant points in the ground-truth boundary of that slice are selected. 

The performance of the full \model{} is compared with the output of the first block in terms of IoU and ASD for different nodule sizes and textures.
As in a real case scenario, we consider that the experts can keep either the initial or the corrected segmentation, according to which better fits their needs. The evaluation is thus performed via Eq.~\ref{eq:eval_2}:

\begin{equation}
\overline{\text{Cr IoU}} = \frac{1}{N}\sum_{n=1}^N \text{max}(\text{IoU}(S_{n,j},\text{Cr}\hat{S}_{n,j}) ,\text{IoU}(S_{n,j},\hat{S}_j))
\label{eq:eval_2}
\end{equation}

This principle is also applied to the ASD metric having as decision criteria the IoU, \textit{i.e.,} the same nodules are considered.

%\subsection{Hyper-parameter selection}
%The \model's hyper-parameters were found via random search, which tends to be better than grid-search for tuning~\cite{Bergstra2011}. We selected $\lambda_1$, $\gamma$ and $p$ that maximized the IoU on the validation set of the first train-test split. At each of the 100 search steps, $\{\lambda_1,\gamma,p\} \sim U([0,1])$, where $U$ is an uniform distribution.    

%% file: results.tex
\section{Results and Discussion}
\label{sec:results}

\begin{figure*}
\centering
\includegraphics[width=1\textwidth]{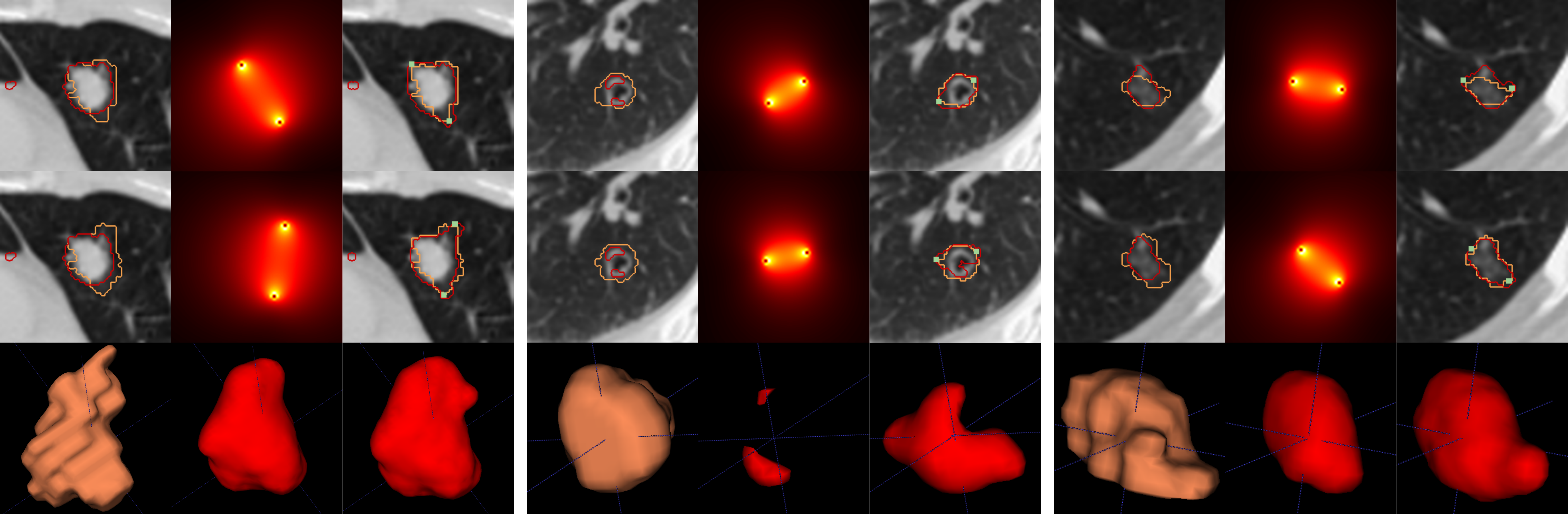}
\caption{Examples of segmentations proposed by \model{}. For each of the $3\times 3$ \protect\includegraphics[width=0.8em,height=0.8em]{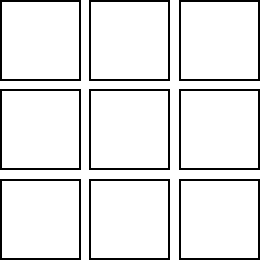} block: 
\protect\includegraphics[width=0.8em,height=0.8em]{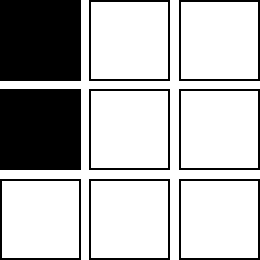} ground-truth ({\color{red}$\blacksquare$}) and output of the first block of \model{} ({\color{orange}$\blacksquare$}) for two different annotators;
\protect\includegraphics[width=0.8em,height=0.8em]{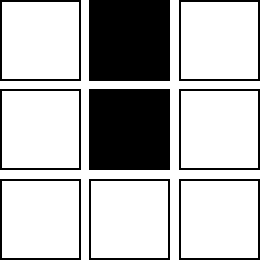} weight maps created based on the end-points of the diameter;
\protect\includegraphics[width=0.8em,height=0.8em]{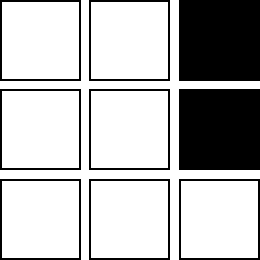} resulting segmentations after considering the diameter's end-points ({\color{green}$\blacksquare$});
\protect\includegraphics[width=0.8em,height=0.8em]{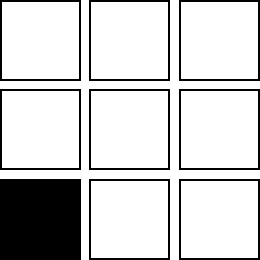} example of a 3D representation of the ground-truth from the nodule above;
\protect\includegraphics[width=0.8em,height=0.8em]{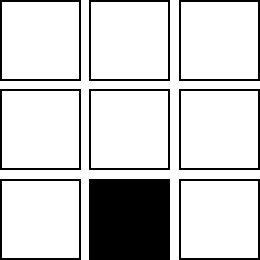}
3D representation of the initial segmentation;
\protect\includegraphics[width=0.8em,height=0.8em]{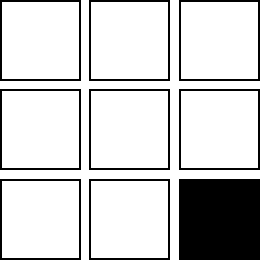}
3D representation of the guided segmentation. \qquad\qquad \label{fig:seg_examples}}

\end{figure*}

\subsection{Hyper-parameters}

The best performing set of parameters are $\gamma=0.59$, $p=0.44$ and $\lambda_1=0.68$. These allow to achieve an average validation IoU of 0.59 in the first train/test split. Intuitively, a $p$ near 0.5 (recall Fig.~\ref{fig:w0.5}) allows to create a weight map that prioritizes the inclusion of the points and the respective connection region without overspreading (Fig.~\ref{fig:w0}) or over-emphasizing the points (Figs.~\ref{fig:w1}~and~\ref{fig:w1}). Likewise, the found $\gamma$ allows the binarized weight map to have an ellipsoidal structure, following the approximate shape of most of the nodules. Finally, $\lambda_1$ balances the contribution of the initial manual segmentation and the added weight map during model training. In the limit where $\lambda_1=0$ the network would be trying to approximate the nodule segmentation to an ellipsoid. On the other hand, $\lambda_1=0.68$ ensures that the manual segmentation is the prioritized target during training and that the weight map \M{} (see Fig.~\ref{fig:seg_examples}) is used for local corrections.

\subsection{Experiment 1}

\begin{figure}
\centering
\includegraphics[width=0.9\columnwidth]{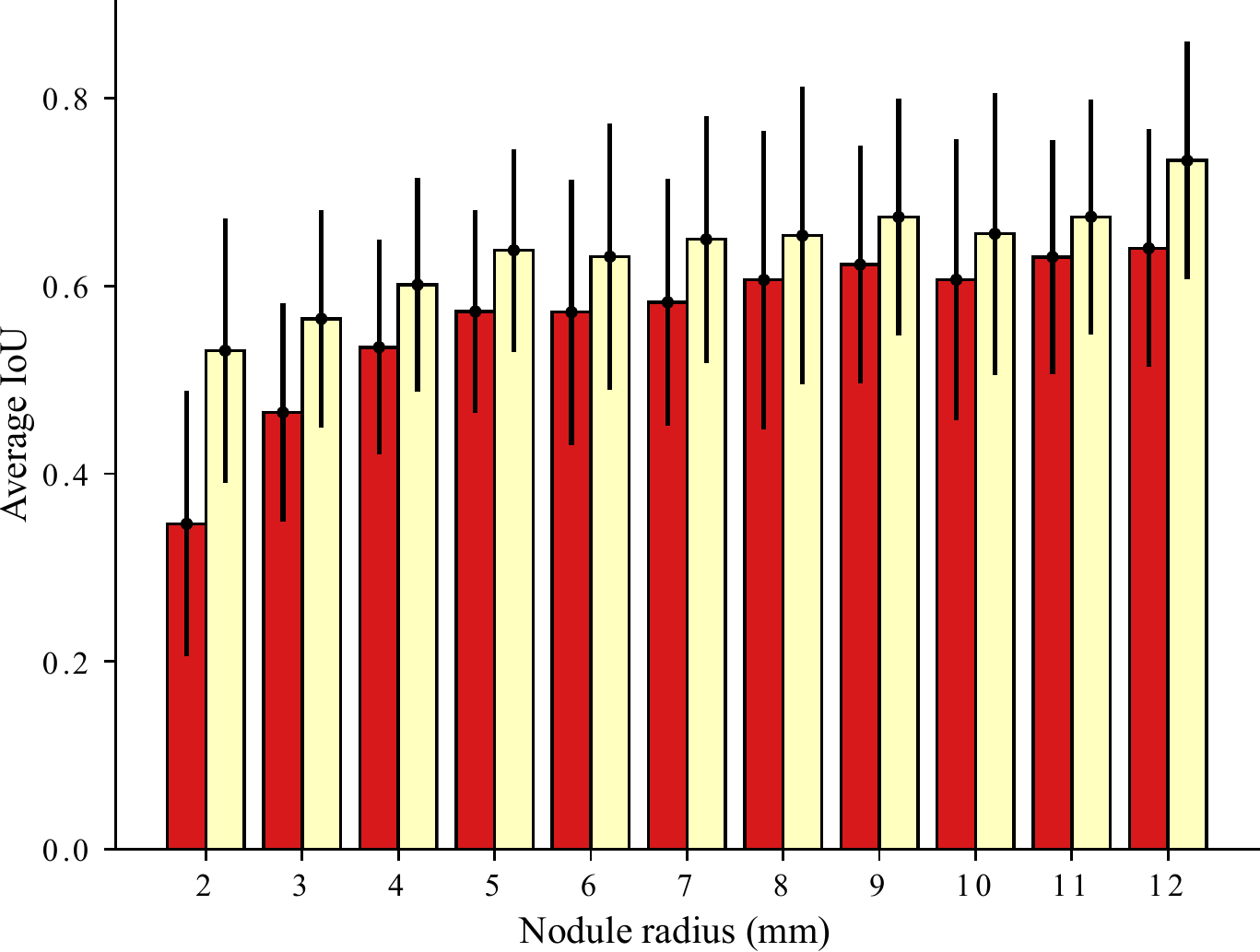}
\caption{Average Intersection over Union per nodule radius for the initial segmentation of \model{} ({\color{red}$\blacksquare$}) and the inter-observer agreement ({\color{yellow}$\blacksquare$}), and the respective standard deviation. \label{fig:rad}}
\end{figure}

\model{} without user interaction outperforms the baseline 3D \U{}~\cite{10.1007/978-3-319-46723-8_49}. As shown in Table~\ref{tab:exp1}, the nodule segmentation performance is relatively increased by approximately 39\% while reducing the number of parameters by a factor of 6.9. In fact, the reduction of the size of the network contributed to the disparity between the referred IoUs by allowing to increase the batch size during training and thus help the error's back-propagation via a better batch normalization~\cite{Ioffe2015}.

As expected, \model{}'s prediction without user-interaction tends to be better for larger nodules (see Fig.~\ref{fig:rad}). Indeed, since most segmentation errors occur near the nodules' boundary, then smaller nodules, which have a higher surface area \textit{vs} volume ratio, should be more challenging. Interestingly, the inter-observer agreement follows the same tendency, indicating that smaller nodules are particularly difficult to segment.

\begin{table}
\centering
\caption{Intersection over Union $\pm$ the standard deviation of the prediction of the first block \model{} in comparison to a 3D \U{} and the inter-observer agreement. \label{tab:exp1}}
\begin{tabular*}{\columnwidth}{c @{\extracolsep{\fill}}cc}\hline
&\textbf{IoU}& \textbf{Number of parameters} \\ \hline
Inter-observer & $0.59\pm0.14$ & -\\
3D \U{}~\cite{10.1007/978-3-319-46723-8_49}  & $0.38\pm0.08$ & 19 080 001 \\
\model{} first block & $0.48\pm0.19$ & 1 592 093 \\
\hline %3 184 722
\end{tabular*}
\end{table}

\subsection{Experiment 2}

\begin{table}
\centering
\caption{Percentage of the number of improved segmentations and respective Average absolute intersection over union increase (IoU improv.) $\pm$ the standard deviation of \model{}'s guided segmentation in comparison to the initial segmentation.\label{tab:improvement}}
\begin{tabular*}{\columnwidth}{l @{\extracolsep{\fill}}cccc} \hline
\textbf{Nodule type} & \textbf{All} & \textbf{Solid} & \textbf{Sub-solid} & \textbf{Non-solid} \\ \hline
Improv. (\%) & 78 & 78 & 73 & 87 \\
IoU improv. & $0.08\pm0.10$ & $0.07\pm0.08$ & $0.08\pm0.09$ & $0.15\pm0.13$ \\\hline

\end{tabular*}

\end{table}

\begin{figure}
\centering
\includegraphics[width=0.9\columnwidth]{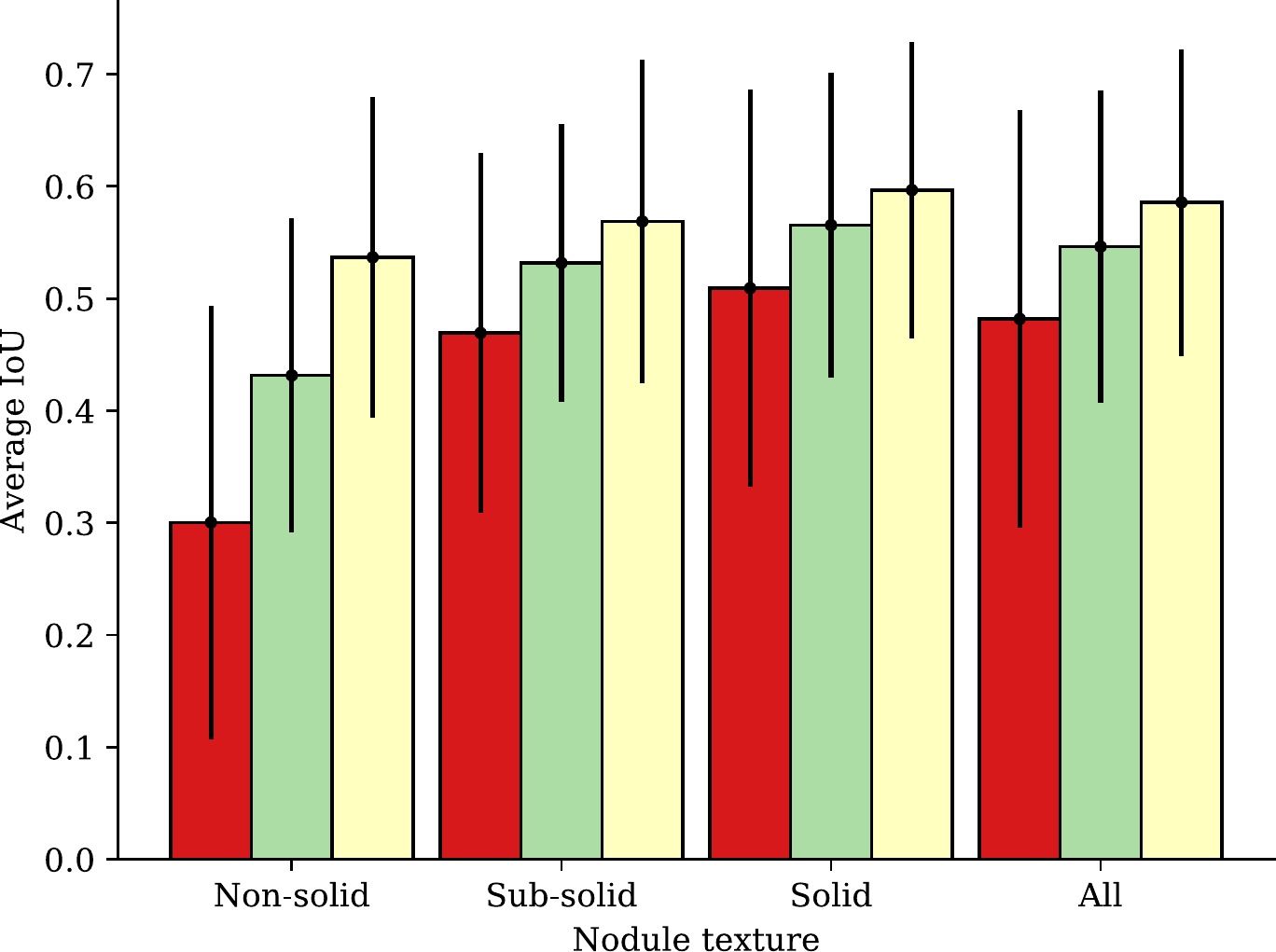}
\caption{Average Intersection over Union per nodule texture for \model{}'s initial ({\color{red}$\blacksquare$}) and corrected segmentations ({\color{green}$\blacksquare$}), the inter-observer agreement ({\color{yellow}$\blacksquare$}), and the respective standard deviation.\label{fig:texture}}
\end{figure}

The proposed simplistic user interaction approach allows to improve the baseline segmentation on more than 75\% of the cases.  Fig.~\ref{fig:seg_examples} depicts examples where \model{} allows to significantly alter the 3D shape of the segmentation just by the introduction of two points, being capable of correcting, at least partially, poor segmentations (middle) as well as change the orientation of the proposed region of interest (right). In fact, 44\% of the user-introduced points are inside the new segmentations, further showing the tendency of \model{} to alter the shape of the segmentation.
Also, as detailed in Table~\ref{tab:improvement} and Fig.~\ref{fig:texture}, \model{} specially enables the delineation correction of the challenging non-solid nodules. 

\begin{figure}
\centering
\includegraphics[width=0.9\columnwidth]{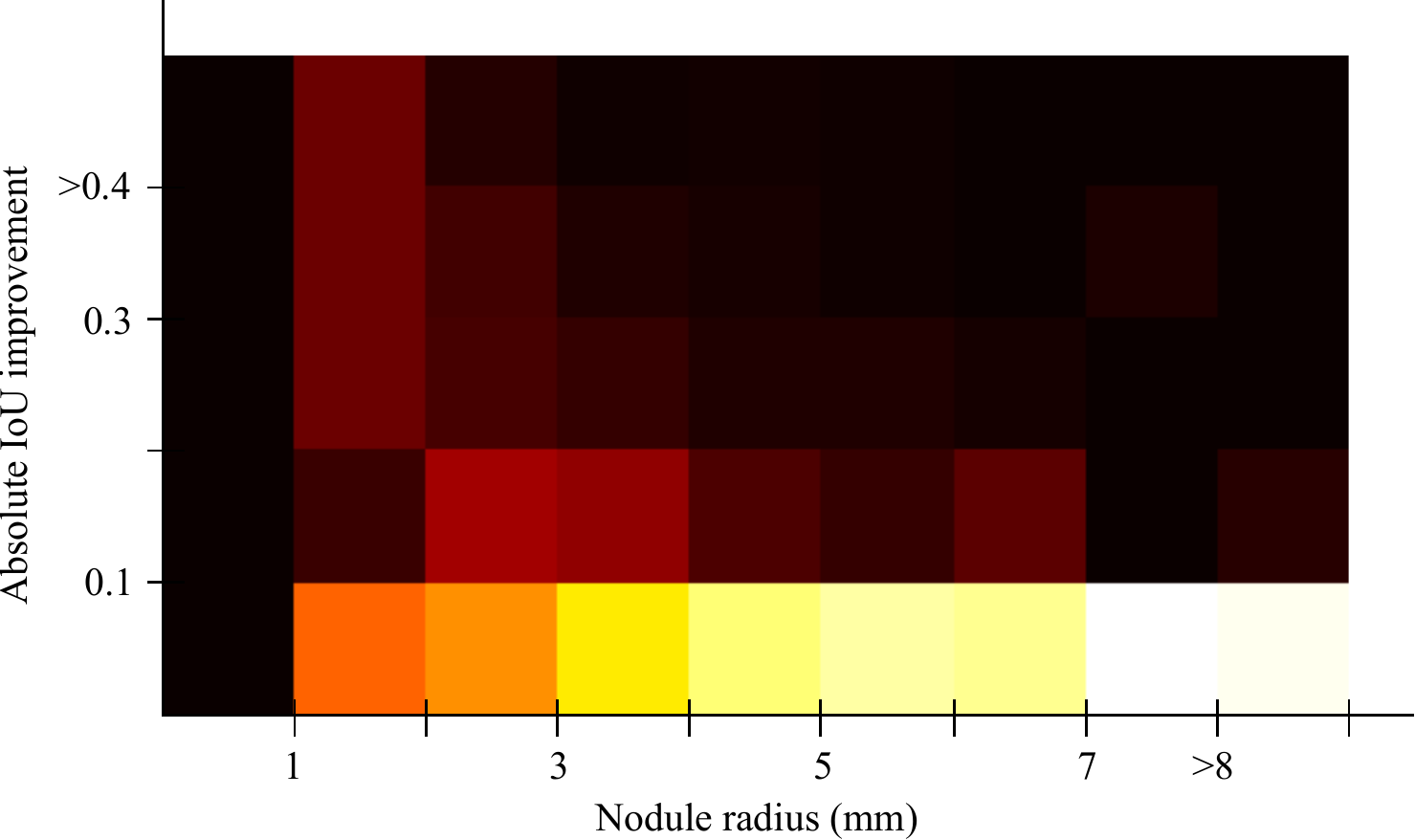}
\caption{Average absolute Intersection over Union improvement between the initial and the corrected segmentation using \model{} per nodule radius. Each column is normalized according to the respective number of nodules.  
%Only nodules with agreement level 2 and higher are considered.  
Colorbar: 0~\protect\includegraphics[width=4em,height=0.5em]{colorbar}~1.0\label{fig:iou_guided}}
\end{figure}

Our proposed approach also has promising results for computer-aided lung cancer screening. As depicted in Fig.~\ref{fig:iou_guided}, the radius range $[1,4](mm)$ is where \model{} (user supervised) most improves the quality of the nodules' segmentation. Importantly, several international lung cancer screening guidelines, such as LUNG-RADS~\cite{ACR2014}, point this dimension range as essential to classify a nodule as either benign or malignant. 

\model{} with the simulated user-interaction allows to improve over the baseline for nodules of different dimensions and textures, as summarized in Fig.~\ref{fig:seg_examples},~\ref{fig:texture}~and~\ref{fig:iou_guided}.
However, the achieved IoU is still, in average, 0.04 lower than the inter-observer agreement. A possible reason for this is that, due to the variability of the ground-truth in the data (\textit{i.e.} several segmentations for the same nodule), the network is likely to learn an average segmentation in order to minimize the loss over the redundant training images. Also, during the segmentation correction we are always selecting the two furtherest points in the nodule boundary. In fact, this is a challenging scenario since there is no guarantee that the selected points are in the direction in which the segmentation needs to be corrected. Instead, we are assuming that providing an estimation of the nodule's largest axis is sufficient to improve the segmentation.

\begin{figure}[h]
\centering
\includegraphics[width=0.9\columnwidth]{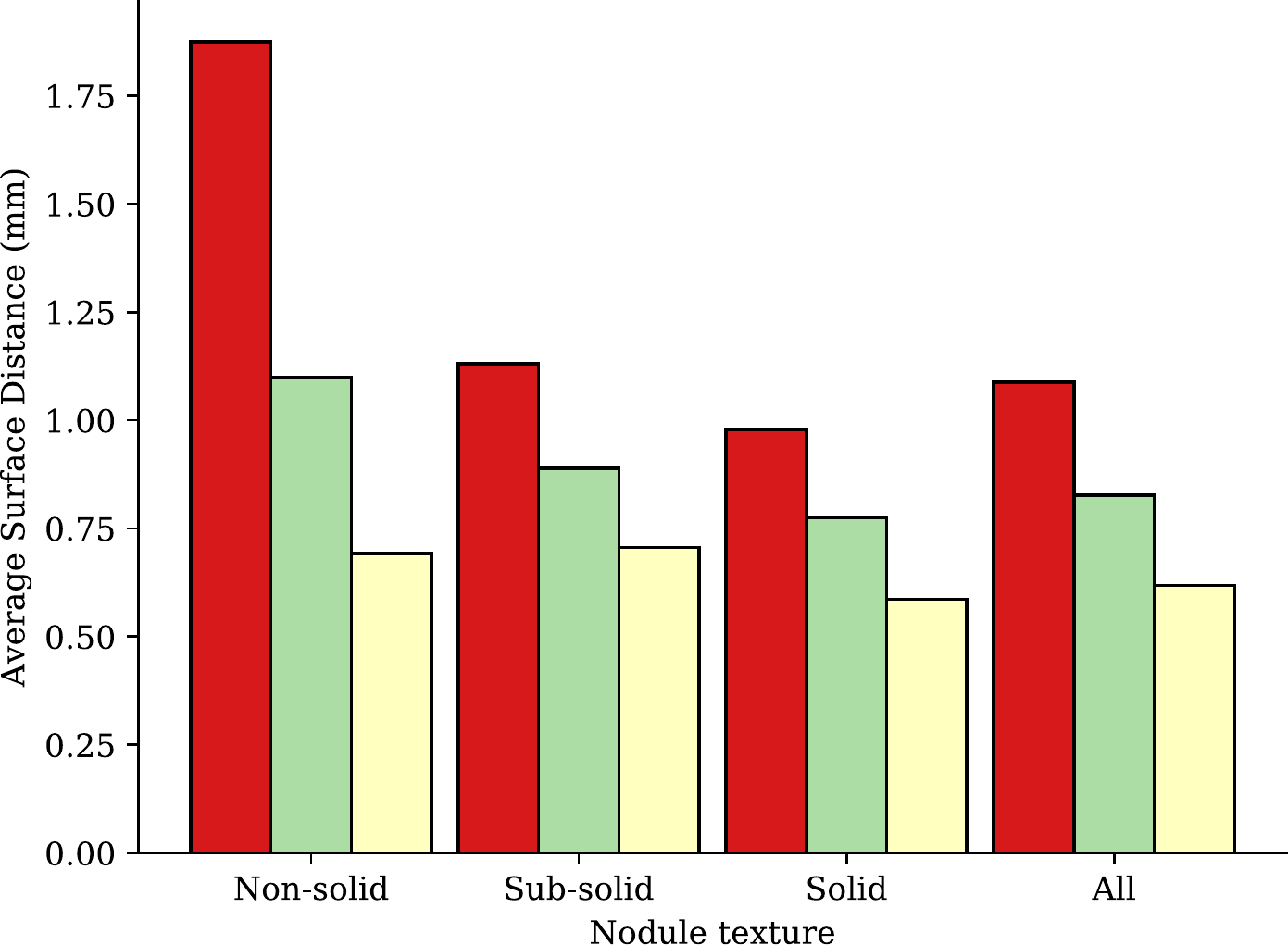}
\caption{Average surface distance (ASD) per nodule texture using \model{} for the initial segmentation ({\color{red}$\blacksquare$}), corrected segmentation ({\color{green}$\blacksquare$}) and the inter-observer agreement ({\color{yellow}$\blacksquare$}).\label{fig:asd}}
\end{figure}

%ASD
% inter-observer 
% corrected 0.827
% original 1.09
% inter-observer 0.62
Despite always using the two farthest points to correct the segmentation, \model{} improves the baseline segmentation's ASD by 24\%, (Fig.~\ref{fig:asd}). Namely, the baseline's average ASD is 1.09 and the corrected's is 0.827, meaning that \model{} has a segmentation error that is in average less than 1 voxel. Also, similarly to the IoU's behavior, the simplistic user interaction allows to significantly improve the quality of the nodules' segmentation in non-solid and sub-solid abnormalities. 

\subsection{Comparison with other approaches}

\model{} achieves a performance in pair to the inter-observer agreement, similarly to other state-of-the-art approaches. 
Note that making a direct comparison between the approaches is non-trivial since
\begin{inparaenum}
\item there is a great variation on the size of the test set, type and size of the nodules used as well as the minimum inter-observer agreement;
\item different methods use different voxel scales, and the inherent re-sampling affects the shape of the ground-truth;
\item there are different ways of combining the ground-truth annotations from the different observers (using all, the average or the median, for instance) to produce the final evaluation mask.
\end{inparaenum}
Nevertheless, for reference, Table~\ref{tab:sota} shows the achieved IoUs of different approaches on the LIDC-IDRI dataset. Similarly to other state-of-the-art approaches, the performance of our method is close to the inter-observer agreement, even though a significantly larger number of samples has been studied. Advantageously, \model{} does not rely on computationally heavy pre-processing steps and allows to segment nodules of all sizes and textures without the need to define bounding boxes or other specific parameters. 
Also, unlike \etal{Wu}~\cite{Wu2018} model, training \model{} does not require other metadata, making it easier to enrich the training set and thus the generalization capability of the system.

\begin{table}
\centering
\caption{Average intersection over union $\pm$ the standard deviation for lung nodule segmentation methods on the LIDC-IDRI dataset, and the reported inter-observer agreement (Inter). NA: information is not available. *Sub-solid nodules only. \label{tab:sota}}
\begin{tabular*}{\columnwidth}{l @{\extracolsep{\fill}}ccccc}
\hline
\textbf{Approach} & \textbf{Year} & \multicolumn{2}{c}{\textbf{\# Nodules}} & \textbf{IoU} & \textbf{Inter} \\
 & & \textbf{Train} & \textbf{Test} & & \\ \hline
\etal{Tan}~\cite{Tan2013b} & 2013 & NA & 23 & 0.65 & NA \\
\etal{Lassen}~\cite{Lassen2015}* & 2015 & NA & 19 & $0.52\pm0.07$ & $0.54\pm0.05$  \\
\etal{Messay}~\cite{Messay201548} & 2015 & 300 & 66 & $0.74\pm 0.11$ & NA \\
\etal{Gon\c{c}alves}~\cite{Goncalves20161} & 2016 & 57 & 512 & $0.71\pm 0.07$ & $0.71\pm0.1$ \\
\etal{Wang}~\cite{Wang2017} & 2017 & 350 & 493 & $0.71\pm 0.12$ & $0.72\pm 0.04$ \\
\etal{Wu}~\cite{Wu2018} & 2018 & 1404 & 1404 & $0.58\pm 0.02$ & NA \\ \hline
\model{} & 2018 & 1593 & 1593 & $0.55\pm 0.14$ & $0.59\pm 0.14$ \\ \hline

\end{tabular*}

\end{table}

%% file: conclusions.tex
\section{Conclusion}
\label{sec:conclusions}

We propose \model{}, a novel lung nodule interactive segmentation scheme. Drawing a stroke of the nodule's diameter and respective end-point extraction allows to generate a weight map \M{}, which is then used for altering the prediction of the network. Specifically, \M{} is designed having in account the expected spherical shape of the nodules and the distance between the introduced points. To promote the influence of \M{} in the resulting segmentation, this map is incorporated as a feature of the model and as a component of the loss function.

\model{} allows to improve the segmentation of more than 75\% of the studied nodules. In fact, in comparison to the baseline, our model (with user interaction) significantly improves the segmentation of nodules with radii $[1,4](mm)$, which are essential for referral. Likewise, using \model{} improves the segmentation performance of nodules with all types of textures, specially the challenging non-solid nodules.
Given the inherent subjectivity of lung nodule segmentation, \model{} may be an important tool to add to CAD systems, removing the need for manual segmentation while providing an easy and fast method to correct the produced output if needed.